\documentclass[lettersize,journal]{IEEEtran}
\usepackage{amsmath}
\usepackage{amsfonts}
\usepackage{algorithmic}
\usepackage{array}
\usepackage[T1]{fontenc}
\usepackage[table]{xcolor}
\usepackage[caption=false,font=normalsize,labelfont=sf,textfont=sf]{subfig}
\usepackage{textcomp}
\usepackage{stfloats}
\usepackage{url}
\usepackage{verbatim}
\usepackage{graphicx}
\usepackage{cite}
\usepackage{balance}
\usepackage{booktabs} 
\usepackage{enumitem} 
\usepackage{makecell}

\usepackage{multirow}
\usepackage[pagebackref,breaklinks,colorlinks,citecolor=cvprblue,urlcolor=cvprblue,linkcolor=cvprblue]{hyperref}

\usepackage{colortbl}
\usepackage{longtable}
\definecolor{Gray}{gray}{0.93}
\definecolor{Red}{RGB}{255, 46, 23}
\usepackage[ruled,vlined]{algorithm2e} 
\SetKwComment{Comment}{\tiny// }{}
\usepackage{amssymb} 
\usepackage{pifont}
\usepackage{tikz}
\usetikzlibrary{positioning, arrows.meta, decorations.pathreplacing}
\definecolor{my_deep_yellow}{cmyk}{0, 0.008, 0.1, 0.08}
\definecolor{my_yellow}{cmyk}{0, 0.002, 0.04, 0.020}
\definecolor{cvprblue}{rgb}{0.21,0.49,0.74}
\definecolor{my_blue}{cmyk}{0.04, 0.02, 0, 0}
\definecolor{my_green}{cmyk}{0.02, 0, 0.04, 0.012}
\definecolor{acl_green}{cmyk}{0.74,0,0.21,0.40}
\definecolor{mycyan}{cmyk}{0.065, 0, 0, 0}
\usepackage{orcidlink} 

\begin{document}

\title{Parameter- and Bandwidth-Efficient Edge--cloud Many-to-Many Speech-to-Text Translation}

\author{Yexing Du\orcidlink{0009-0003-0513-2635}, Kaiyuan Liu\orcidlink{0000-0001-7359-4450}, Youcheng Pan\orcidlink{0000-0002-8270-5455}, Bo Yang\orcidlink{0000-0002-4288-8349},\\Lei Chen\orcidlink{0009-0005-4334-2873}, Ming Liu\orcidlink{0000-0001-7915-1001}, Bing Qin\orcidlink{0000-0002-2543-5604}, Yang Xiang\orcidlink{0000-0003-1395-6805}
\thanks{Manuscript received xxx 2026; This work was supported in part by Harbin Institute of Technology and Pengcheng Laboratory. (\textit{Corresponding authors: Ming Liu; Yang Xiang.}) }%
\thanks{Yexing Du, and Kaiyuan Liu are with Harbin Institute of Technology, Shenzhen, China, and also with Pengcheng Laboratory, Shenzhen, China (e-mail: yxdu@ir.hit.edu.cn; 1171000408@stu.hit.edu.cn).

Youcheng Pan, Bo Yang, and Yang Xiang are with Pengcheng Laboratory, Shenzhen, China (e-mail: panych@pcl.ac.cn; yangb05@pcl.ac.cn; xiangy@pcl.ac.cn).

Lei Chen is with the Bay Area International Business School, Beijing Normal University, Beijing, China (e-mail: chenlei@bnu.edu.cn).

Ming Liu and Bing Qin are with Harbin Institute of Technology, Harbin, China, and also with Pengcheng Laboratory, Shenzhen, China (e-mail: mliu@ir.hit.edu.cn; qinb@ir.hit.edu.cn).

}%
}

\markboth{Journal of \LaTeX\ Class Files,~Vol.~14, No.~8, August~2021}%
{Shell \MakeLowercase{\textit{et al.}}: A Sample Article Using IEEEtran.cls for IEEE Journals}

\maketitle

\begin{abstract}
Multimodal large language models (MLLMs) have demonstrated significant potential for speech-to-text translation (S2TT). However, existing deployment paradigms face critical challenges: pure on-device models suffer from resource constraints, while centralized cloud systems incur bandwidth bottlenecks and privacy risks by transmitting raw voice data. In this paper, we propose Edge--cloud Speech Recognition and Translation (ESRT), a parameter-efficient, bandwidth-efficient, and privacy-aware collaborative Edge--cloud MLLM framework. First, we introduce a multi-task weighted curriculum learning strategy to mitigate catastrophic forgetting, improve multilingual balance, and train parameter-efficient ESRT-1B, ESRT-4B, and ESRT-12B models. Second, we enable bandwidth-efficient Edge--cloud inference by retaining a lightweight speech encoder and adapter on the device and transmitting only a compressed tensor to the cloud. Extensive experiments on FLEURS demonstrate that ESRT models achieve state-of-the-art S2TT performance across 45 languages ($45 \times 44$ directions). Relative to raw audio, ESRT and ESRT-Lite reduce the transmitted tensor size by $5.1\times$ and $10.2\times$, respectively, while keeping raw speech on-device and avoiding its direct exposure to the cloud. The code and models are released to facilitate reproducible, privacy-aware S2TT research.
\end{abstract}


\section{Introduction}
Speech-to-text translation (S2TT) converts source-language speech into target-language text. Traditional systems cascade automatic speech recognition (ASR) and machine translation (MT), making them susceptible to error propagation~\cite{sperber2020speech}. Recently, multimodal large language models (MLLMs) have unified speech understanding and text generation within a simpler architecture. Their deployment, however, faces resource, bandwidth, and privacy constraints: edge devices lack sufficient computation and memory, while cloud inference requires bandwidth-intensive raw-audio streaming that exposes voiceprint biometrics.

Existing speech services therefore face a fundamental deployment trade-off. They typically follow one of two paradigms: (1) \textbf{Offline Edge Inference}: Limited computation, memory, and energy budgets force the use of compact on-device models. Although local processing avoids network dependence and keeps audio on the device, the reduced model capacity often limits translation quality,. (2) \textbf{Online Cloud Inference}: Cloud models provide stronger multilingual modeling, but they require users to upload raw audio. This design incurs substantial communication overhead, and directly exposes raw audio to the cloud service.

\begin{figure}[t]
\centering
 \includegraphics[width=\linewidth]{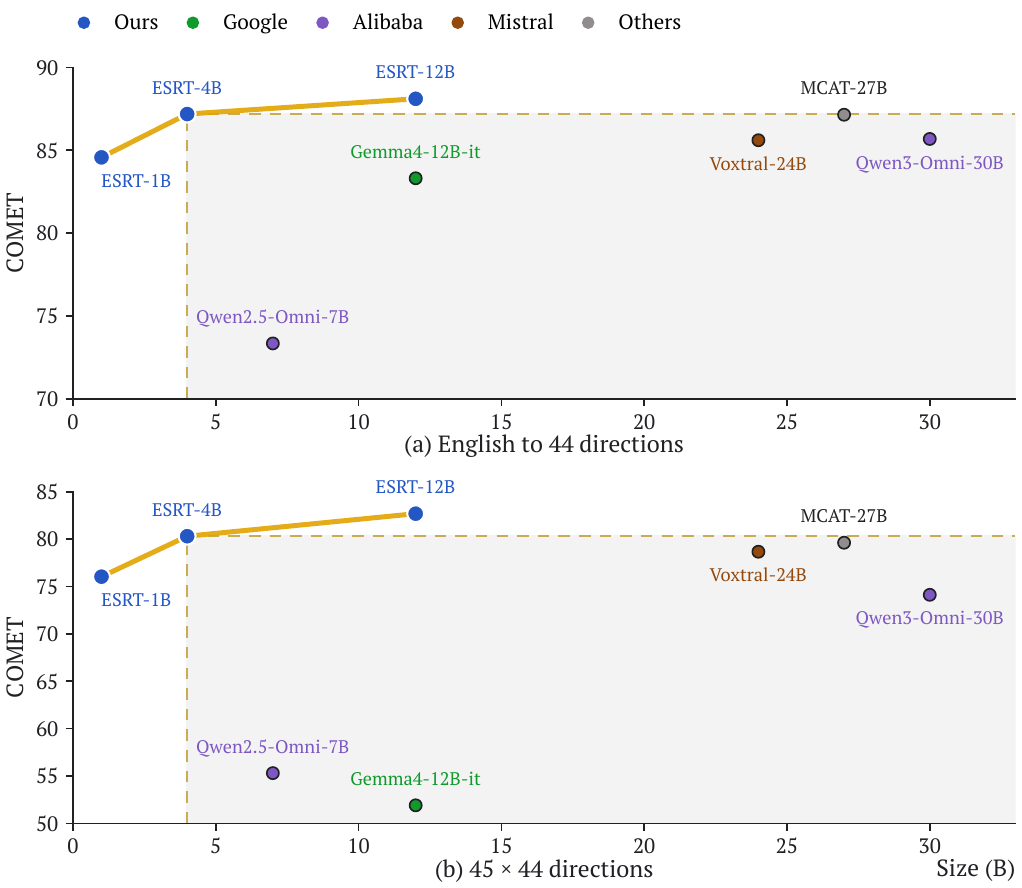}
\caption{\textbf{Model scale vs. multilingual S2TT performance.} Existing MLLMs typically rely on larger parameter counts to support multilingual capabilities. ESRT demonstrates strong parameter efficiency, achieving superior multilingual performance with substantially fewer parameters.}
\label{fig:mul}
\end{figure}

ESRT addresses these limitations through two designs. (1) \textbf{Parameter Efficiency by Multi-Task Weighted Curriculum Learning}: We extend LLM-SRT~\cite{du2025making} with weighted task designs to activate multilingual S2TT. As shown in Figure~\ref{fig:mul}, our approach outperforms substantially larger models with fewer parameters. (2) \textbf{Bandwidth Efficiency by Edge--cloud Inference}: We retain the speech encoder and adapter on the edge and transmit only a compressed tensor. Avoiding raw-audio uploads reduces communication overhead and direct exposure of the original waveform. We further reuse the transmitted tensor, eliminating repeated transfers.

\clearpage

\begin{figure}[t]
\centering
 \includegraphics[width=1\linewidth]{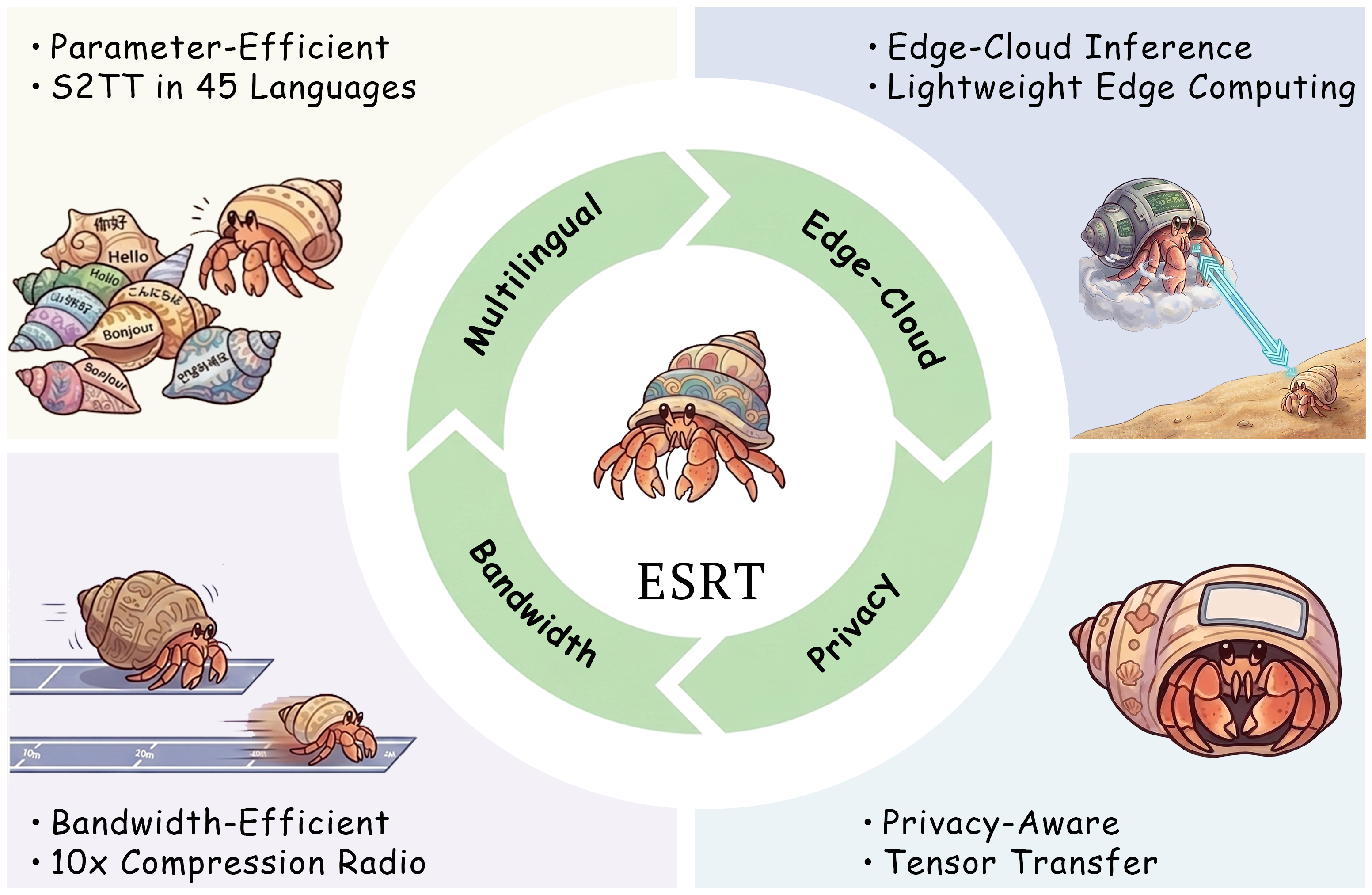}
\caption{\textbf{Overview of the ESRT features.} (1) \textbf{Multilingual Parameter Efficiency}: ESRT models support many-to-many S2TT in 45 languages. (2) \textbf{Edge--cloud Inference}: The speech encoder and adapter run on the edge, while the LLM handles high-capacity inference in the cloud. (3) \textbf{Bandwidth Efficiency}: Compared with raw speech, the compressed tensor achieves an approximately $10\times$ size reduction. (4) \textbf{Privacy-Aware Data Minimization}: Raw speech remains on the device, reducing direct audio exposure. }
\label{fig:intro}
\end{figure}

As shown in Figure~\ref{fig:intro}, ESRT offers four key capabilities. 
We evaluate \textbf{ESRT-1B}, \textbf{ESRT-4B}, and \textbf{ESRT-12B}, each using 80 speech tokens, under the complete 45-language protocol ($45\times44$ directions) on FLEURS~\cite{conneau2023fleurs}. ESRT-4B and ESRT-12B outperform strong end-to-end and cascaded baselines. Comparing \textbf{ESRT-12B} (80 speech tokens) with \textbf{ESRT-12B-Lite} (40 speech tokens) shows that the latter retains competitive translation quality while achieving a $\mathbf{10.2\times}$ raw-audio-to-representation size reduction. Our analyses additionally cover 45 languages, representation size and memory efficiency on heterogeneous hardware, and ablation studies.

Our main contributions are summarized as follows:
\begin{itemize}
    \item We introduce an \textbf{Edge--cloud speech-to-text translation framework} that maintains high translation quality while reducing raw-data transmission and avoiding direct upload of the original audio.
    \item We introduce a \textbf{multi-stage weighted curriculum learning} strategy and release parameter-efficient ESRT-1B, ESRT-4B, and ESRT-12B models for many-to-many S2TT.
    \item We conduct a \textbf{systematic evaluation of MLLMs for many-to-many S2TT} over all $45\times44$ translation directions, revealing their performance characteristics across speech encoder, resource levels, model scales, and deployment settings.
    \item The framework supports \textbf{both training and inference on NPUs and GPUs}.\footnote{The code and models are released at \url{https://github.com/yxduir/esrt}.}
\end{itemize}

Departing from the curriculum-based adaptation of LLM-SRT~\cite{du2025making} and the multilingual scaling of MCAT~\cite{du2026mcat}, this study pivots to resource-constrained edge scenarios. We enhance parameter efficiency via a multi-stage weighting strategy and propose an edge--cloud framework that transmits intermediate tensors rather than raw audio, simultaneously reducing bandwidth and protecting data privacy.

\newpage
\section{Related Work}\label{sec:related}

\subsection{Speech-to-Text Translation}

Most large-scale S2TT research and corpora have historically been \emph{English-centric}, exemplified by datasets like CoVoST-2~\cite{wang2020covost} and cascaded systems that pivot through English text (e.g., Whisper~\cite{radford2023robust} paired with MT~\cite{nllb2024scaling}). This approach simplifies data collection but often leads to under-supervised non-English targets and compounding errors. Recent MLLMs can inherit this bias, showing strong performance for $\text{English} \rightarrow X$ directions but struggling with fully non-English directions.

MLLMs integrate ASR, translation, and understanding into a single autoregressive decoder, promising to mitigate cascaded error propagation~\cite{sperber2020speech}. Representative models include SALMONN~\cite{tang2023salmonn} with a dual-encoder architecture connected to an LLM via a Q-Former, and SpeechGPT~\cite{zhang2023speechgpt} which extends an LLM with speech capabilities through cross-modal instruction tuning. 
However, these MLLM-based S2TT systems are inherently monolithic. The entire model, including the speech encoder and the LLM, must be deployed on a single device, which limits their use in resource-constrained edge scenarios.

Many-to-many S2TT directly supports a broad grid of source and target languages without an English pivot. SeamlessM4T-V2-Large unifies multilingual speech and text tasks in an encoder--decoder framework and covers extensive translation directions~\cite{seamless2025joint}. LLM-SRT introduces a staged curriculum of speech recognition, machine translation, and speech translation to adapt LLMs for multilingual S2TT~\cite{du2025making}. MCAT subsequently scales MLLM-based many-to-many S2TT to 70 languages, demonstrating the benefit of increased model capacity and multilingual supervision~\cite{du2026mcat}. 

\subsection{Edge--cloud Computing}
Edge--cloud computing~\cite{liang2020ai-on-edge} has emerged as a critical paradigm for deploying AI models, particularly for latency-sensitive and privacy-aware applications like S2TT. This approach offloads computationally intensive tasks to the cloud while keeping sensitive data processing closer to the user on edge devices. Prior works have explored various architectures for distributing AI inference: Auto-Split~\cite{autosplit} proposes dynamic DNN partitioning based on network conditions and device capabilities; CoEdge~\cite{coedge} enables collaborative inference across heterogeneous edge devices; ED-ViT~\cite{liu2024edvit} identifies optimal split points within Vision Transformer self-attention blocks; and Splitwise~\cite{younesi2025splitwise} uses a DRL policy to adapt fine-grained LLM partitioning under variable edge--cloud conditions. The key challenge in Edge--cloud systems is deciding where to split the computation pipeline~\cite{zhang2025}.

Speech introduces distinct deployment requirements: raw waveforms are temporally dense and costly to transmit, intermediate features can remain large, and an aggressive compression or poorly chosen split point may discard acoustic information needed for recognition and translation.  An effective split-inference design for S2TT must therefore jointly balance edge-side resource consumption, representation size, translation quality, and raw-data exposure.

\clearpage
\begin{figure*}[t]
\centering
 \includegraphics[width=\linewidth]{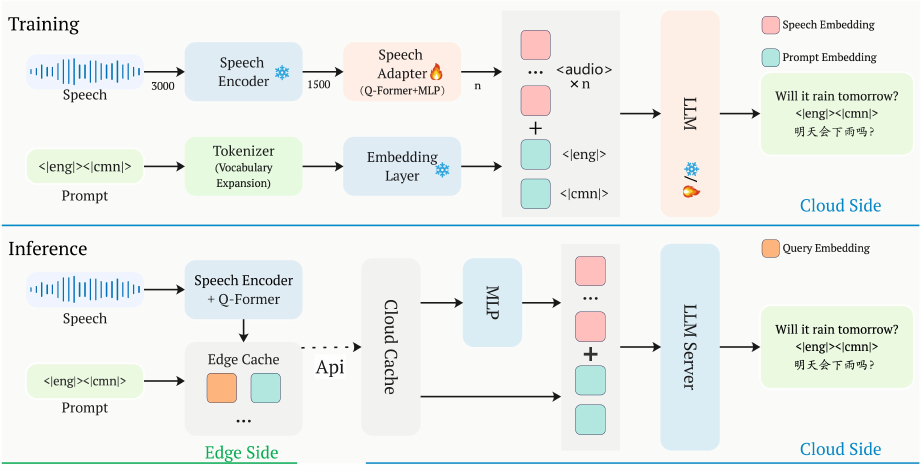}
\caption{\textbf{ESRT architecture.} The framework implements collaborative inference across the edge and cloud through an API that transmits a compressed tensor rather than raw speech, substantially reducing communication traffic. By eliminating raw waveform transmission, this design also narrows the attack surface, since the cloud receives only the intermediate tensor required for inference.}
\label{fig:framework}
\end{figure*}

\begin{table}[t] 
\centering 
\small
\caption{MLLM Training Settings.} 
\renewcommand{\arraystretch}{1.2} 
\setlength{\tabcolsep}{10pt} 
\resizebox{1.0\linewidth}{!}{
\begin{tabular}{lcl} 
\toprule
\textbf{Modules} & \textbf{Train Stage} & \textbf{Details} \\
\midrule
Speech Encoder & - & Whisper's encoder \\
\midrule
Speech Adapter & All & Q-Former / MLP \\
\midrule
LLM & - & MiLMMT 1B/4B/12B \\
\hspace{1em}- Vision Encoder  &  -  &  Removing visual encoder                  \\
\hspace{1em}+ Vocabulary & - & Replace 102 <unused> tokens  \\ 
\hspace{1em}+ Lora & III & r=16, alpha=32 \\ 
\bottomrule
\end{tabular}}
\label{tab:parameters} 
\end{table}

\section{Methodology}
\subsection{MLLM Architecture}
As detailed in Figure \ref{fig:framework} and Table \ref{tab:parameters}, \textbf{ESRT} adopts an MLLM architecture consisting of a pre-trained LLM, a frozen Whisper encoder~\cite{radford2023robust}, and a Q-Former-based~\cite{li2023blip} speech adapter~\cite{du2025making}. 

\label{sec:3.1}
\subsubsection{\textbf{Task Formulation}}
\label{sec:3.0}
\begin{itemize}[itemsep=0.3ex,topsep=0.4ex]
\item \textbf{Automatic Speech Recognition (ASR):} Given speech \(x\) and instruction \(t\), generate its transcription \(Y_1\).
\item \textbf{Speech-guided Machine Translation (SMT):} Given \(x\), \(t\), and \(Y_1\), generate the translation \(Y_2\).
\item \textbf{Speech Recognition and Translation (SRT):} Given \(x\) and \(t\), jointly generate \(Y_1\) and \(Y_2\).
\end{itemize}

\subsubsection{\textbf{Backbone Adaptation}}
ESRT-1B/4B/12B use the corresponding MiLMMT variants~\cite{shang2026scalingmodeldatamultilingual}, which are derived from Gemma-3~\cite{team2025gemma}. We remove the unused vision encoder and repurpose 102 \texttt{<unused>} tokens as FLEURS language identifiers (e.g., \texttt{<|eng|>} and \texttt{<|cmn|>})~\cite{conneau2023fleurs} to control the target language and reduce language switching.

\begin{table}[t]
  \centering
  \small
  \renewcommand{\arraystretch}{1.23} 
  \caption{Prompt Design.}
  \setlength{\tabcolsep}{2pt}
  \resizebox{1.0\linewidth}{!}{
    \begin{tabular}{c c c l l} 
    \toprule 
    \textbf{Stage} & \textbf{Weight} & \textbf{Task} & \textbf{Prompt} & \textbf{Prediction} \\ 
    \midrule

    I & 1.0 & ASR & {<|eng|>} & \{Text\} \\ 
    \midrule

    & 0.2 & ASR & {<|eng|>} & \{Text\} \\ 
    
    II & 0.4 & SMT & {}\{Text\}{<|eng|><|deu|>} & \{Translation\} \\ 
   & 0.4 & SRT & {<|eng|><|deu|>} & \{Text\}{<|eng|><|deu|>}\{Translation\} \\ 
    \midrule

    \multirow{2}{*}{III} & 0.2 & ASR & {<|eng|>} & \{Text\} \\ 
    & 0.8 & SRT & {<|eng|><|deu|>} & \{Text\}{<|eng|><|deu|>}\{Translation\} \\ 

    \bottomrule
    \end{tabular}}
\label{tab:prompt_design} 
\end{table}

\subsubsection{\textbf{Multi-Task Weighted Curriculum Learning Strategy}}
LLM-SRT~\cite{du2025making} adopts a three-stage curriculum that trains ASR, SMT, and SRT sequentially. Although this progression gradually introduces translation ability, switching completely to the later tasks can weaken the ASR capability acquired in Stage~I. This degradation is particularly relevant to SRT because its autoregressive target first generates the source-language transcription and then the target-language translation. Recognition errors in the first part of the output can therefore propagate to the subsequent translation and reduce the final S2TT score. To preserve acoustic grounding throughout the curriculum, we replace the strictly sequential task schedule with the weighted mixtures in Table~\ref{tab:prompt_design}. The weights specify task-sampling proportions within each stage. Stage~I uses ASR to establish speech--text alignment; Stage~II samples ASR/SMT/SRT at 0.2/0.4/0.4 to introduce translation while maintaining recognition; and Stage~III samples ASR/SRT at 0.2/0.8 to emphasize direct S2TT while continuing ASR supervision.

\clearpage

\begin{figure}[t]
\centering
 \includegraphics[width=\linewidth]{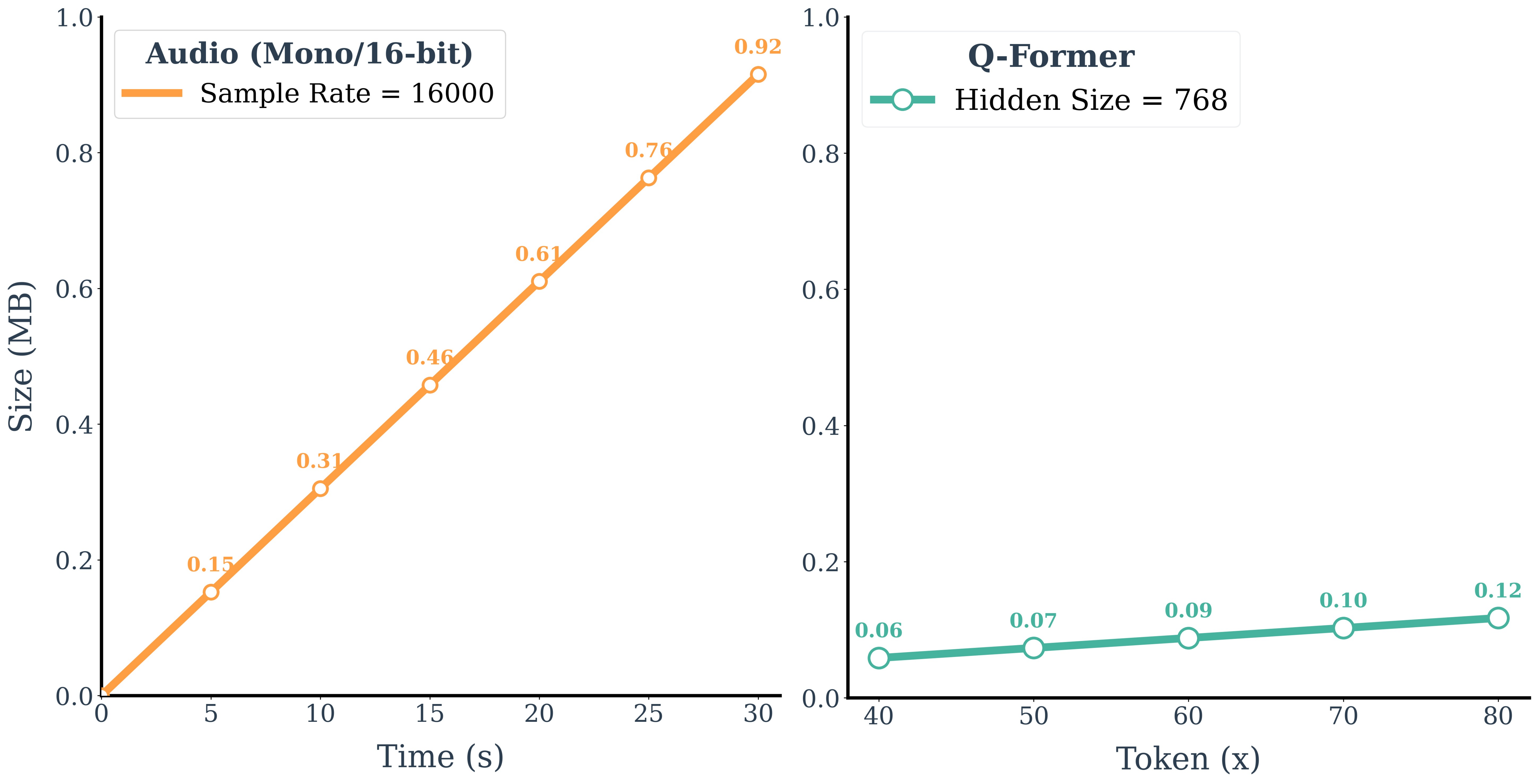}
\caption{\textbf{Data size analysis.} We compare raw-audio size with Q-Former tensor size under different token budgets; the following analysis uses 40 tokens.}
\label{fig:size}
\end{figure}

\subsection{\textbf{Edge--cloud Inference}}

\paragraph{\textbf{Bandwidth Analysis: Tensor vs.\ Audio Size}}
Figure~\ref{fig:size} compares raw WAV audio ($S_{\text{wav}}$) with its compressed tensor ($S_{\text{tensor}}$). For 30-second mono audio sampled at 16~kHz with 16-bit depth ($B_d=2$):
\begin{equation}
    S_{\text{wav}} = T \times f_s \times B_d = 960,000\text{ bytes} \ (\approx 0.92\text{ MB}).
\end{equation}

For the 40-token configuration, a BF16 Q-Former tensor with $L_c=40$, $D_{\text{q}}=768$, and $B_t=2$ bytes per element occupies
\begin{equation}
    S_{\text{tensor}} = L_c \times D_{\text{q}} \times B_t = 61,440\text{ bytes} \ (\approx 0.059\text{ MB}).
\end{equation}
Because the Q-Former produces a fixed number of speech tokens, the tensor payload has a fixed size for inputs within the 30-second window. The break-even duration relative to direct transmission of 16-kHz, 16-bit mono audio is
\begin{equation}
    T_{\text{break-even}} = \frac{S_{\text{tensor}}}{f_s \times B_d}
    = \frac{61,440}{16,000 \times 2} = 1.92\text{ s}.
\end{equation}
Thus, for audio longer than approximately 1.92 seconds, transmitting the tensor requires a smaller payload than transmitting the raw waveform, and the reduction increases with audio duration.

\paragraph{\textbf{Privacy-Aware Inference}}
Inspired by shared-state and memory-reuse mechanisms in multi-agent~\cite{liu2026projectevalplus} systems, ESRT further supports \textbf{Feature Cache} when the same speech input requires multiple inference requests, such as translation into multiple target languages. The compressed tensor is transmitted only for the first request and subsequently reused from the cloud cache, avoiding repeated transmission of either the raw speech or its intermediate representation. 

ESRT follows a data-minimization design: the original waveform is processed on the edge and is not uploaded to the cloud. Instead, the Q-Former compresses the speech encoder output into a fixed number of learned speech tokens. For the Lite configuration, the transmitted tensor contains 40 speech tokens rather than the raw audio samples or the full frame-level encoder sequence. Inputs are also padded to a fixed 30-second window, producing a fixed-shape speech-token representation. These design choices reduce direct exposure of the original waveform and make straightforward waveform recovery less direct than when raw audio is transmitted. 
\section{Experimental Settings}

\paragraph{\textbf{Datasets}}
In our experiments, we employ a curriculum learning strategy across distinct training phases. Specifically, we utilize the \textit{CommonVoice 24} dataset in Stage I. Subsequently, the \textit{FLEURS} dataset~\cite{conneau2023fleurs} is introduced for training during Stage II and Stage III. Finally, we conduct comparative and ablation studies on both the \textit{FLEURS} and \textit{CoVoST-2}~\cite{wang2020covost} datasets.

\paragraph{\textbf{Language Support}}
As shown in Table~\ref{tab:language_support_summary}, ESRT supports 45 languages across high-, medium-, and low-resource groups, covering all $45\times44=1{,}980$ non-identity translation directions. The training corpus contains \textbf{398.5} hours of S2TT data in total, with fewer than 10 hours per language.

\paragraph{\textbf{Training Details}}
As shown in Table~\ref{tab:parameters}, the proposed MLLM comprises an LLM (MiLMMT~\cite{shang2026scalingmodeldatamultilingual}), a frozen speech encoder, and a trainable speech adapter. The standard ESRT models use 80 speech tokens, whereas ESRT-Lite uses 40.
All reported ESRT checkpoints are trained on $8\times$ Ascend 910C NPUs using BF16 precision and DeepSpeed Zero-0. The same training pipeline also supports NVIDIA GPUs and yields consistent performance. We use AdamW with a learning rate of $5 \times 10^{-5}$ and a 1000-step linear warmup schedule.

\paragraph{\textbf{Compared Methods}}

\begin{itemize}
    \item \textbf{API Models}: We evaluate the commercial Gemini-3.5-Flash-Lite API.
    \item \textbf{Cascade Systems}: We implement a pipeline model as a strong traditional baseline. Specifically, we employ Whisper-Large-V3~\cite{radford2023robust} as the ASR module and NLLB-3.3B~\cite{nllb2024scaling} as the downstream MT model.
    \item \textbf{End-to-End Models}: We compare against SeamlessM4T-V2-Large~\cite{seamless2025joint}, MCAT-27B~\cite{du2026mcat}, ZeroSwot-Large~\cite{tsiamas-etal-2024-pushing}, Gemma4-12B-it~\cite{gemmateam2026gemma4}, Voxtral-24B~\cite{liu2025voxtral}, Qwen2.5-Omni-7B~\cite{Qwen2.5-Omni}, and Qwen3-Omni-30B~\cite{xu2025qwen3}.
\end{itemize}

\paragraph{\textbf{Evaluation Metrics}} 
We employ COMET and spBLEU as our evaluation metrics. Specifically, spBLEU utilizes the FLORES-200 tokenizer.

\begin{table}[b]
\centering
\small
\renewcommand{\arraystretch}{1}
\setlength{\tabcolsep}{1pt}
\caption{Language support.}
\label{tab:language_support_summary}
\begin{tabular}{@{}l@{\hspace{6pt}}p{0.76\columnwidth}@{}}
\toprule
\textbf{Resource} & \textbf{Languages (ISO Code)} \\
\midrule
High & Arabic (ara), Bengali (ben), Catalan (cat) \\
    & Czech (ces), Chinese (cmn), German (deu) \\
     & English (eng), Finnish (fin), French (fra) \\
     & Italian (ita), Japanese (jpn), Dutch (nld) \\
     & Polish (pol), Romanian (ron), Spanish (spa) \\
\midrule
Medium & Danish (dan), Greek (ell), Hindi (hin) \\
      & Croatian (hrv), Hungarian (hun), Indonesian (ind) \\
       & Kazakh (kaz), Korean (kor), Portuguese (por) \\
       & Russian (rus), Slovak (slk), Tamil (tam) \\
       & Tagalog (tgl), Thai (tha), Turkish (tur) \\
       & Urdu (urd), Uzbek (uzb), Vietnamese (vie) \\
\midrule
Low    & Azerbaijani (azj), Bulgarian (bul), Persian (fas) \\
     & Hebrew (heb), Khmer (khm), Lao (lao) \\
       & Malay (msa), Burmese (mya), Norwegian (nob) \\
       & Slovenian (slv), Swedish (swe), Cantonese (yue) \\
\midrule
\textbf{Total} & \textbf{45 languages} \\
\bottomrule
\end{tabular}
\end{table}

\clearpage

\begingroup
\renewcommand{\arraystretch}{1}
\setlength{\tabcolsep}{6pt}
\begin{table*}[t]
\centering
\scriptsize
\caption{spBLEU / COMET Results for All $45\times44$ Non-Identity Directions on the FLEURS Dataset.}
\resizebox{1.0\textwidth}{!}{
\begin{tabular}{>{\columncolor{gray!8}}l|>{\columncolor{my_green}}c>{\columncolor{my_green}}c>{\columncolor{my_green}}c>{\columncolor{my_green}}c>{\columncolor{my_green}}c>{\columncolor{my_green}}c>{\columncolor{my_green}}c|>{\columncolor{my_green}}c>{\columncolor{my_green}}c>{\columncolor{my_green}}c}
\toprule
& Whisper+ & Gemini-3.5 & SeamlessM4T & Gemma4 & Voxtral & Qwen3-Omni & MCAT & ESRT-1B & ESRT-4B & ESRT-12B \\
\multirow{-2}{*}{\textbf{Directions}} & NLLB-3.3B & Flash-Lite & V2-Large & 12B-it & 24B & 30B-Instruct & -27B & (ours) & (ours) & (ours) \\
\midrule
ara$\rightarrow$44 & 21.1 / 78.7 & \underline{22.2} / \underline{80.2} & 15.7 / 70.1 & 17.4 / 74.4 & 20.1 / 79.8 & 20.5 / 78.8 & 20.0 / 79.7 & 15.6 / 75.8 & 20.8 / 80.0 & \textbf{25.8} / \textbf{83.2} \\
azj$\rightarrow$44 & \underline{16.9} / 78.2 & 14.7 / 75.6 & 13.6 / 73.5 & 1.1 / 41.0 & 15.3 / 77.6 & 7.7 / 65.2 & 15.6 / 78.8 & 11.7 / 75.0 & 16.2 / \underline{79.8} & \textbf{20.2} / \textbf{82.8} \\
ben$\rightarrow$44 & 8.0 / 60.5 & \textbf{20.6} / \textbf{82.3} & 14.6 / 73.9 & 1.5 / 43.3 & 12.1 / 72.9 & 11.6 / 70.9 & 14.8 / 76.6 & 10.2 / 71.0 & 15.0 / 76.8 & \underline{19.9} / \underline{81.2} \\
bul$\rightarrow$44 & 22.8 / 80.1 & 20.1 / 75.7 & 18.6 / 73.9 & 3.3 / 46.8 & 23.0 / 81.2 & 18.4 / 75.5 & 23.1 / \underline{82.0} & 17.6 / 77.7 & \underline{23.4} / \underline{82.0} & \textbf{27.1} / \textbf{84.1} \\
cat$\rightarrow$44 & 26.2 / 83.0 & \underline{27.8} / 83.3 & 22.2 / 77.9 & 10.4 / 60.6 & 26.1 / 83.4 & 21.7 / 78.3 & 24.9 / 83.7 & 20.8 / 81.4 & 26.0 / \underline{84.5} & \textbf{29.4} / \textbf{86.0} \\
ces$\rightarrow$44 & 22.5 / 81.0 & 20.7 / 77.5 & 18.1 / 73.7 & 2.2 / 44.4 & 22.5 / 82.0 & 19.1 / 77.2 & 22.7 / 82.8 & 17.7 / 78.9 & \underline{23.4} / \underline{83.2} & \textbf{26.6} / \textbf{84.8} \\
cmn$\rightarrow$44 & 18.5 / 80.5 & 17.2 / 79.5 & 13.4 / 74.0 & 0.7 / 43.3 & 16.7 / 80.6 & \underline{20.9} / \underline{82.8} & 17.7 / 81.3 & 14.1 / 78.4 & 18.0 / 81.3 & \textbf{21.3} / \textbf{83.3} \\
dan$\rightarrow$44 & 23.4 / 80.8 & 18.6 / 73.9 & 17.6 / 73.4 & 1.5 / 36.9 & \underline{24.3} / 82.4 & 22.4 / 80.0 & 23.3 / 82.5 & 18.3 / 78.6 & 23.9 / \underline{82.8} & \textbf{27.8} / \textbf{84.9} \\
deu$\rightarrow$44 & 25.8 / 83.4 & \underline{27.6} / 83.8 & 19.3 / 74.9 & 22.8 / 79.8 & 26.7 / 84.0 & 26.9 / 83.8 & 25.2 / 83.9 & 20.4 / 81.2 & 25.7 / \underline{84.4} & \textbf{28.9} / \textbf{85.7} \\
ell$\rightarrow$44 & \underline{21.1} / 80.1 & 13.5 / 67.7 & 14.3 / 71.1 & 0.7 / 36.6 & 20.0 / 79.5 & 12.9 / 69.3 & 20.1 / 80.2 & 14.7 / 75.9 & 20.5 / \underline{80.7} & \textbf{24.4} / \textbf{83.0} \\
eng$\rightarrow$44 & 30.5 / 84.3 & \underline{34.2} / 86.6 & 31.8 / 85.3 & 29.9 / 83.3 & 31.5 / 85.6 & 31.9 / 85.7 & 31.9 / 87.1 & 26.9 / 84.6 & 32.5 / \underline{87.2} & \textbf{35.4} / \textbf{88.1} \\
fas$\rightarrow$44 & 19.1 / 77.8 & \underline{22.6} / 80.6 & 14.9 / 69.9 & 0.7 / 39.8 & 16.4 / 76.1 & 14.2 / 71.6 & 19.1 / 80.1 & 13.9 / 75.3 & 19.5 / \underline{80.7} & \textbf{24.5} / \textbf{83.9} \\
fin$\rightarrow$44 & \underline{21.9} / 82.7 & 16.4 / 75.5 & 13.9 / 69.5 & 0.6 / 37.7 & 18.5 / 79.7 & 17.8 / 78.5 & 21.2 / 82.9 & 15.3 / 78.0 & 21.7 / \underline{83.4} & \textbf{25.4} / \textbf{85.2} \\
fra$\rightarrow$44 & 24.9 / 83.6 & \underline{27.0} / 83.6 & 19.2 / 77.0 & 22.1 / 79.8 & 25.7 / 83.8 & 25.9 / 83.8 & 24.4 / 84.1 & 20.1 / 81.3 & 24.8 / \underline{84.3} & \textbf{28.1} / \textbf{85.8} \\
heb$\rightarrow$44 & \underline{18.7} / 73.2 & 12.9 / 62.6 & 15.6 / 67.6 & 0.6 / 35.4 & 15.7 / \underline{73.7} & 7.4 / 56.6 & 15.3 / 72.7 & 11.7 / 67.6 & 16.3 / 73.3 & \textbf{22.2} / \textbf{78.2} \\
hin$\rightarrow$44 & 17.9 / 78.4 & \underline{23.7} / \underline{83.5} & 13.2 / 71.9 & 15.7 / 74.7 & 19.4 / 81.7 & 20.0 / 81.4 & 21.9 / 83.3 & 15.0 / 78.5 & 21.1 / 83.1 & \textbf{24.9} / \textbf{84.9} \\
hrv$\rightarrow$44 & 22.1 / 80.8 & 19.6 / 76.7 & 16.6 / 70.3 & 3.3 / 46.4 & 22.0 / 81.4 & 15.7 / 75.9 & 23.2 / 83.0 & 17.6 / 78.7 & \underline{23.3} / \underline{83.2} & \textbf{26.7} / \textbf{84.9} \\
hun$\rightarrow$44 & \underline{21.2} / 79.9 & 14.8 / 70.6 & 15.3 / 68.9 & 0.5 / 36.7 & 18.4 / 78.7 & 11.0 / 65.0 & 19.6 / 79.9 & 14.9 / 75.7 & 20.6 / \underline{80.8} & \textbf{24.8} / \textbf{83.7} \\
ind$\rightarrow$44 & 23.7 / 82.2 & \underline{27.4} / \underline{84.5} & 15.2 / 71.5 & 2.1 / 46.6 & 23.4 / 83.2 & 24.7 / 83.4 & 24.2 / 84.2 & 18.9 / 80.5 & 24.7 / 84.4 & \textbf{27.6} / \textbf{85.5} \\
ita$\rightarrow$44 & 22.7 / 83.1 & \underline{25.6} / 84.4 & 17.1 / 75.8 & 20.9 / 81.3 & 24.0 / 83.8 & 23.3 / 83.4 & 22.5 / 83.8 & 19.0 / 81.6 & 23.4 / \underline{84.6} & \textbf{26.2} / \textbf{85.8} \\
jpn$\rightarrow$44 & 18.9 / 80.5 & 18.4 / 80.8 & 11.9 / 69.5 & 13.5 / 74.5 & 17.9 / 81.2 & \underline{19.5} / \underline{81.8} & 18.0 / 80.8 & 13.7 / 77.6 & 17.7 / 80.9 & \textbf{21.4} / \textbf{83.1} \\
kaz$\rightarrow$44 & 15.2 / 73.5 & \underline{16.1} / \underline{76.0} & 12.5 / 70.0 & 1.0 / 39.3 & 8.9 / 66.4 & 3.0 / 47.8 & 12.4 / 72.0 & 9.2 / 67.2 & 14.6 / 75.0 & \textbf{20.1} / \textbf{80.5} \\
khm$\rightarrow$44 & 0.0 / 28.8 & \textbf{11.1} / \textbf{71.1} & 8.8 / 65.6 & 0.3 / 39.9 & 6.3 / 65.0 & 1.1 / 43.3 & 4.8 / 61.1 & 4.4 / 60.7 & 7.0 / 65.6 & \underline{10.5} / \underline{69.7} \\
kor$\rightarrow$44 & 19.5 / 81.8 & 20.8 / 83.1 & 14.2 / 74.0 & 13.8 / 74.4 & 18.0 / 81.2 & 20.4 / 82.7 & \underline{20.9} / \underline{83.4} & 15.1 / 79.2 & 20.2 / 83.3 & \textbf{23.7} / \textbf{85.0} \\
lao$\rightarrow$44 & 6.3 / 59.9 & \underline{14.0} / \underline{73.4} & 10.6 / 67.3 & 0.4 / 40.6 & 9.8 / 70.0 & 10.2 / 66.9 & 8.3 / 67.8 & 6.6 / 65.1 & 10.6 / 70.8 & \textbf{14.4} / \textbf{74.2} \\
msa$\rightarrow$44 & 22.1 / 80.4 & \underline{23.6} / 81.2 & 12.9 / 67.3 & 2.2 / 46.7 & 22.0 / 81.7 & 21.0 / 79.7 & 22.9 / 82.5 & 17.3 / 78.6 & 23.1 / \underline{83.0} & \textbf{26.9} / \textbf{85.0} \\
mya$\rightarrow$44 & 0.1 / 28.0 & \underline{6.1} / \underline{65.1} & \textbf{8.9} / \textbf{65.6} & 0.2 / 37.1 & 1.5 / 49.7 & 0.9 / 43.2 & 1.4 / 49.9 & 1.2 / 50.1 & 2.8 / 53.7 & 4.0 / 55.4 \\
nld$\rightarrow$44 & 22.1 / 82.3 & 21.6 / 79.9 & 16.5 / 73.7 & 7.9 / 56.9 & \underline{22.4} / 82.8 & 22.0 / 82.2 & 22.2 / 83.3 & 16.9 / 79.6 & 21.9 / \underline{83.4} & \textbf{24.9} / \textbf{84.9} \\
nob$\rightarrow$44 & 22.3 / 80.4 & 22.6 / 78.8 & 16.5 / 72.3 & 2.1 / 40.5 & 24.0 / 82.5 & 23.0 / 81.2 & 24.1 / \underline{83.6} & 19.1 / 79.9 & \underline{24.2} / 83.5 & \textbf{27.7} / \textbf{85.3} \\
pol$\rightarrow$44 & 21.0 / 81.2 & 19.8 / 78.5 & 13.9 / 69.9 & 3.0 / 47.3 & 21.5 / 82.0 & 20.0 / 80.2 & \underline{21.7} / \underline{82.9} & 16.4 / 79.2 & 21.3 / \underline{82.9} & \textbf{24.8} / \textbf{84.6} \\
por$\rightarrow$44 & 26.2 / 83.5 & \textbf{30.9} / \underline{85.6} & 16.3 / 71.8 & 24.9 / 81.1 & 27.5 / 84.4 & 27.3 / 83.8 & 26.5 / 85.0 & 21.9 / 82.7 & 26.9 / 85.4 & \underline{29.9} / \textbf{86.4} \\
ron$\rightarrow$44 & 24.8 / 81.9 & 23.9 / 79.8 & 18.9 / 73.1 & 6.4 / 55.0 & 24.9 / 82.7 & 22.5 / 79.8 & 24.8 / 83.3 & 19.4 / 79.8 & \underline{25.2} / \underline{83.8} & \textbf{28.5} / \textbf{85.4} \\
rus$\rightarrow$44 & 23.7 / 82.8 & \underline{26.7} / \underline{83.8} & 18.3 / 75.3 & 21.3 / 79.4 & 24.6 / 83.0 & 24.4 / 82.6 & 24.1 / \underline{83.8} & 19.4 / 80.9 & 23.8 / 83.7 & \textbf{27.3} / \textbf{85.2} \\
slk$\rightarrow$44 & 23.3 / 82.0 & 22.6 / 79.9 & 17.9 / 73.2 & 3.8 / 48.2 & 22.6 / 81.6 & 17.5 / 76.6 & 23.4 / 83.1 & 18.2 / 79.3 & \underline{23.9} / \underline{83.5} & \textbf{27.6} / \textbf{85.3} \\
slv$\rightarrow$44 & 19.6 / 77.7 & 15.1 / 71.2 & 14.8 / 68.0 & 2.1 / 41.2 & 18.4 / 78.2 & 11.9 / 67.0 & 19.7 / 79.6 & 14.5 / 74.9 & \underline{19.9} / \underline{79.7} & \textbf{23.8} / \textbf{82.9} \\
spa$\rightarrow$44 & 22.8 / 83.4 & \textbf{26.1} / \underline{85.3} & 15.9 / 75.2 & 21.0 / 81.2 & 24.1 / 84.2 & 23.0 / 83.3 & 23.1 / 84.6 & 18.8 / 82.0 & 23.1 / 84.8 & \underline{25.8} / \textbf{85.8} \\
swe$\rightarrow$44 & 24.5 / 82.8 & 20.9 / 77.4 & 17.6 / 72.8 & 2.1 / 41.5 & 24.2 / 82.8 & 22.0 / 79.6 & 25.3 / \underline{84.6} & 19.5 / 80.4 & \underline{25.4} / 84.3 & \textbf{28.7} / \textbf{86.0} \\
tam$\rightarrow$44 & 12.9 / 71.4 & \underline{17.3} / \textbf{78.9} & 12.6 / 69.8 & 0.5 / 35.4 & 10.1 / 70.6 & 2.5 / 50.5 & 13.0 / 74.0 & 9.0 / 69.0 & 13.1 / 74.2 & \textbf{17.6} / \underline{78.5} \\
tgl$\rightarrow$44 & 21.9 / 79.4 & \underline{22.4} / 79.1 & 11.5 / 60.2 & 1.6 / 41.5 & 21.4 / 79.2 & 17.9 / 74.6 & 18.6 / 78.1 & 15.2 / 74.4 & 20.5 / \underline{79.8} & \textbf{24.9} / \textbf{82.2} \\
tha$\rightarrow$44 & 17.1 / 78.6 & \underline{20.2} / \underline{82.0} & 11.5 / 69.0 & 0.6 / 41.8 & 16.1 / 79.5 & 18.8 / 81.3 & 15.6 / 78.9 & 12.2 / 75.9 & 17.2 / 80.6 & \textbf{21.1} / \textbf{83.0} \\
tur$\rightarrow$44 & 23.4 / 83.0 & \underline{25.6} / 83.7 & 15.8 / 72.1 & 1.8 / 45.1 & 20.8 / 80.8 & 23.2 / 82.5 & 24.0 / \underline{84.1} & 17.4 / 79.8 & 23.5 / 84.0 & \textbf{27.1} / \textbf{85.8} \\
urd$\rightarrow$44 & 16.2 / 75.3 & \underline{19.6} / \underline{80.4} & 11.3 / 68.2 & 8.0 / 62.1 & 16.1 / 77.5 & 11.0 / 68.3 & 17.3 / 79.2 & 12.2 / 74.4 & 17.4 / 79.8 & \textbf{22.6} / \textbf{83.0} \\
uzb$\rightarrow$44 & 5.6 / 53.9 & 14.1 / \underline{72.4} & \underline{14.7} / 70.5 & 0.7 / 38.3 & 8.5 / 65.1 & 2.9 / 46.6 & 10.1 / 66.7 & 8.2 / 64.5 & 11.9 / 69.6 & \textbf{17.6} / \textbf{75.9} \\
vie$\rightarrow$44 & 18.7 / 78.9 & 19.7 / 80.1 & 14.1 / 72.6 & 0.2 / 37.5 & 17.6 / 79.8 & \underline{20.0} / \underline{81.3} & 18.0 / 80.5 & 14.3 / 77.2 & 18.5 / 81.1 & \textbf{21.9} / \textbf{82.9} \\
yue$\rightarrow$44 & 16.8 / 78.6 & 7.9 / 67.8 & 10.9 / 67.9 & 0.2 / 40.3 & 13.6 / 77.3 & \textbf{19.1} / \textbf{81.8} & 14.0 / 77.6 & 10.8 / 73.8 & 14.3 / 77.5 & \underline{17.5} / \underline{79.9} \\
\midrule
\textbf{Avg.} & 19.4 / 76.5 & \underline{20.3} / 78.3 & 15.3 / 71.5 & 6.6 / 51.9 & 19.2 / 78.7 & 17.3 / 74.1 & 19.7 / 79.6 & 15.2 / 76.0 & 20.2 / \underline{80.3} & \textbf{23.9} / \textbf{82.7} \\
\bottomrule
\end{tabular}
}
\raggedright{\hspace*{1em}\textbf{Bold} values indicate the best results, while \underline{underlined} values denote the second-best results.}
\label{comet_1144}
\end{table*}
\endgroup

\section{Experiments}
\subsection{Many-to-Many S2TT on 45$\times$44 Directions}
\subsubsection{\textbf{Overall Performance}}
Table~\ref{comet_1144} shows that ESRT-12B achieves the best overall performance across all $45\times44$ directions, reaching \textbf{23.9} spBLEU and \textbf{82.7} COMET. Compared with the API-based Gemini-3.5-Flash-Lite, it improves spBLEU by 3.6 points and COMET by 4.4 points. It also exceeds the strongest prior COMET average, MCAT-27B at 79.6, by 3.1 points, demonstrating robust multilingual translation quality beyond selected language subsets.

\subsubsection{\textbf{Performance Scaling}}
ESRT exhibits consistent gains as model capacity increases. The 1B, 4B, and 12B variants obtain 15.2/76.0, 20.2/80.3, and 23.9/82.7 in spBLEU/COMET, respectively. Scaling from 1B to 12B therefore yields improvements of 8.7 spBLEU and 6.7 COMET. The monotonic trend across all 1,980 directions indicates that greater capacity improves both lexical accuracy and semantic adequacy while reducing cross-lingual interference.

\subsubsection{\textbf{Mitigation of Error Accumulation}}
Cascaded S2TT systems compound recognition errors when the source language is poorly represented by the speech encoder. This effect is particularly evident for Burmese (mya) and Khmer (khm): Whisper--NLLB obtains only 28.0 and 28.8 COMET, respectively. Because corrupted acoustic information cannot be recovered by the downstream translation model, these upstream errors propagate through the pipeline. ESRT mitigates this accumulation through end-to-end optimization.

\subsubsection{\textbf{English-Centric vs. Many-to-Many Performance}}
English-centric evaluation can overestimate multilingual robustness. SeamlessM4T-V2-Large performs strongly on English-to-44 translation, reaching 85.3 COMET, but its average falls to 71.5 across the complete many-to-many grid. Similar degradation appears for other general-purpose models when low-resource source languages are included. In contrast, ESRT-12B decreases from 88.1 on English-to-44 directions to 82.7 over all directions, showing better multilingual balance.

\clearpage

 \begin{figure*}[t]
  \centering
   \includegraphics[width=\linewidth]{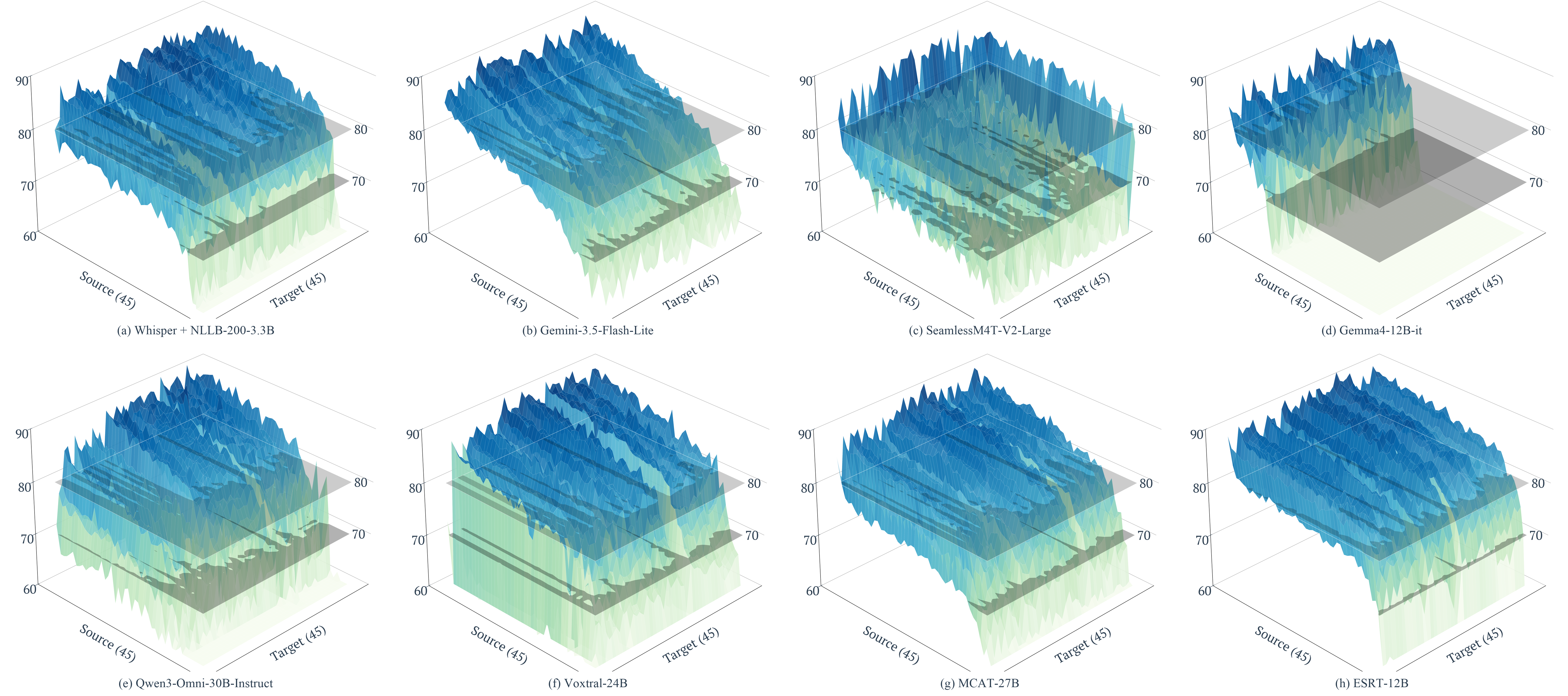}
\caption{\textbf{COMET performance overview for 45 x 45 directions.} Shaded regions highlight scores falling below 80 (light) and 70 (dark). Consequently, the model possessing the minimum total shaded area yields the highest translation performance across the baselines. Identical language pairs, such as eng$\rightarrow$eng on the diagonal, are smoothed for visualization clarity.}
     \label{fig:comet_matrix}
  \end{figure*}

\subsection{Systematic Analysis on $45 \times 44$ Directions}

\begin{table}[h]
\centering
\caption{COMET performance across the $45\times44$ directions.}
\label{tab:model_score_distribution}
\renewcommand{\arraystretch}{1} 
\setlength{\tabcolsep}{6pt} 
\resizebox{1.0\linewidth}{!}{
\begin{tabular}{lcccc}
\toprule
\rowcolor{gray!8}
\textbf{Models} & \textbf{$<70$} & \textbf{$[70,80)$} & \textbf{$\geq80$} & \textbf{$\geq80$ (\%)} \\
\midrule
\rowcolor{my_green}Whisper + NLLB-3.3B  & 262  & 680  & 1038 & 52.4 \\
\rowcolor{my_green}Gemini-3.5-Flash-Lite & 237 & 815  & 928  & 46.9 \\
\rowcolor{my_green}SeamlessM4T-V2-Large & 754  & 1109 & 117  & 5.9 \\
\rowcolor{my_green}Gemma4-12B-it        & 1514 & 256  & 210  & 10.6 \\
\rowcolor{my_green}Qwen3-Omni-30B       & 542  & 557  & 881  & 44.5 \\
\rowcolor{my_green}Voxtral-24B          & 336  & 393  & 1251 & 63.2 \\
\rowcolor{my_green}MCAT-27B             & 194  & 554  & 1232 & 62.2 \\ \midrule
\rowcolor{my_green}ESRT-1B (ours)              & 328  & 1030 & 622  & 31.4 \\
\rowcolor{my_green}ESRT-4B (ours)              & 147  & 525  & 1308 & 66.1 \\
\rowcolor{my_green}ESRT-12B (ours)             & 76   & 339  & \textbf{1565} & \textbf{79.0} \\
\bottomrule
\end{tabular}}
\end{table}

\subsubsection{\textbf{Landscape Analysis}}
Figure~\ref{fig:comet_matrix} visualizes every source--target direction as a point on a COMET performance surface. The unshaded upper region corresponds to $\text{COMET}\geq80$, which generally indicates good translation quality, while the light and dark shaded regions mark directions below $80$ and $70$, respectively. The shaded footprint therefore captures both the frequency and severity of translation failures: isolated patches suggest pair-specific errors, whereas continuous bands or broad blocks reveal systematic language-level weaknesses.

The orientation of each valley further helps localize the underlying limitation. Valleys along the source axis signal weak acoustic encoding across targets, whereas valleys along the target axis expose inadequate generation for a target language. Broad depressions indicate compounded speech-understanding and text-generation weaknesses. Compared with the baselines, \textbf{ESRT-12B} produces the smallest shaded footprint and the fewest deep valleys, demonstrating balanced gains across source and target languages.

\subsubsection{\textbf{Performance Distribution Analysis}}
Table~\ref{tab:model_score_distribution} partitions all $1980$ directions into three quality ranges. From ESRT-1B to ESRT-4B and ESRT-12B, high-quality directions ($\text{COMET}\geq80$) increase from $622$ to $1308$ and $1565$, while severe failures ($\text{COMET}<70$) decrease from $328$ to $147$ and $76$. This distribution shows that scaling ESRT steadily expands reliable language coverage while raising its cross-lingual performance floor.
\subsubsection{\textbf{Analysis of Gemma4-12B-it}}
As shown in Figure~\ref{fig:comet_matrix}, Gemma4-12B-it exhibits a highly uneven landscape: a small subset of directions remains near the high-score surface, whereas broad regions fall below COMET $70$. Table~\ref{tab:model_score_distribution} records $1514$ severe failures and only $210$ high-quality directions. Removing the dedicated speech encoder simplifies the architecture but leaves the model without sufficiently trained acoustic representations for diverse multilingual speech.

\subsubsection{\textbf{Language Coverage of MLLMs}}
As shown in Figure~\ref{fig:comet_matrix}, Qwen3-Omni and Voxtral perform well on several high-resource languages but form recurring valleys for underrepresented sources and targets. Table~\ref{tab:model_score_distribution} reports $542$ and $336$ directions below COMET $70$, respectively. These structured failures show that the scale of a general-purpose MLLM alone cannot ensure balanced many-to-many S2TT when low-resource speech and translation supervision remain limited.

\subsubsection{\textbf{Parameter Efficiency}}
As shown in Figure~\ref{fig:mul}, ESRT achieves a favorable accuracy--size trade-off. ESRT-12B surpasses substantially larger 24B--30B MLLMs, while \textbf{ESRT-4B} outperforms strong models containing approximately six times as many parameters. The consistent advantage of both variants indicates that performance gains arise from the speech-aware architecture and multilingual curriculum rather than brute-force scaling, making ESRT well suited to resource-constrained edge scenarios.

\newpage

\begingroup
\renewcommand{\arraystretch}{1.35}
\setlength{\tabcolsep}{3pt}
\begin{table*}[t]
\centering
\small
\caption{COMET Results on eng $\rightarrow$ 44 Directions on the FLEURS Dataset.}
\resizebox{1.0\textwidth}{!}{
\begin{tabular}{l|ccccccccccccccccccccccc}
\toprule
\rowcolor{gray!8}
& & & & & & & & & & & && & & & & & & & & & &\\
\rowcolor{gray!8}
\multirow{-2}{*}{\textbf{Models}} & \multirow{-2}{*}{ara} & \multirow{-2}{*}{azj} & \multirow{-2}{*}{ben} & \multirow{-2}{*}{bul} & \multirow{-2}{*}{cat} & \multirow{-2}{*}{ces} & \multirow{-2}{*}{cmn} & \multirow{-2}{*}{dan} & \multirow{-2}{*}{deu} & \multirow{-2}{*}{ell} & \multirow{-2}{*}{fas} & \multirow{-2}{*}{fin} & \multirow{-2}{*}{fra} & \multirow{-2}{*}{heb} & \multirow{-2}{*}{hin} & \multirow{-2}{*}{hrv} & \multirow{-2}{*}{hun} & \multirow{-2}{*}{ind} & \multirow{-2}{*}{ita} & \multirow{-2}{*}{jpn} & \multirow{-2}{*}{kaz} & \multirow{-2}{*}{khm} & \\
\midrule
\rowcolor{my_green}Whisper+NLLB-3.3B & 83.9 & 84.0 & 84.3 & 87.7 & 84.1 & 87.3 & 75.3 & 87.7 & 84.4 & 86.2 & 83.7 & 87.9 & 84.7 & 84.5 & 78.1 & 86.7 & 85.6 & 88.6 & 85.2 & 84.9 & 87.4 & 74.9 \\
\rowcolor{my_green}Gemini-3.5-Flash-Lite & 83.8 & 85.8 & 85.1 & 88.9 & 85.3 & 89.1 & 86.4 & 88.7 & 86.2 & 87.1 & 86.5 & 90.9 & 85.8 & 85.2 & \underline{79.0} & 89.0 & \underline{88.1} & 89.8 & 86.3 & 89.7 & 88.6 & 81.4 \\
\rowcolor{my_green}SeamlessM4T-V2-Large & 84.5 & 83.8 & 84.6 & 88.8 & 85.0 & 88.0 & 79.7 & 88.7 & 84.9 & 87.5 & 84.6 & 88.5 & 85.3 & 84.5 & 78.1 & 87.8 & 86.0 & 89.0 & 85.1 & 84.7 & 87.9 & 79.9 \\
\rowcolor{my_green}ZeroSwot-Large & 83.1 & 78.6 & 82.1 & 85.3 & 83.3 & 76.1 & 81.0 & 77.8 & 82.4 & 84.4 & 83.4 & 68.5 & 82.5 & 75.8 & 75.3 & 77.2 & 78.8 & 86.8 & 81.3 & 86.2 & 77.4 & 75.7 \\
\rowcolor{my_green}Qwen2.5-Omni-7B & 84.5 & 61.2 & 63.3 & 82.0 & 81.3 & 82.0 & 86.4 & 82.3 & 85.0 & 74.7 & 73.6 & 76.8 & 84.5 & 65.9 & 59.0 & 77.1 & 75.2 & 86.4 & 84.2 & 88.4 & 50.3 & 38.0 		\\
\rowcolor{my_green}Gemma4-12B-it & 82.3 & 82.5 & 78.7 & 85.6 & 82.6 & 83.9 & 84.6 & 87.6 & 84.4 & 82.6 & 83.8 & 87.5 & 84.5 & 83.0 & 75.7 & 82.9 & 79.0 & 88.2 & 84.6 & 87.6 & 84.8 & 76.0 \\
\rowcolor{my_green}Voxtral-24B & 85.9 & 85.3 & 84.3 & 89.6 & \textbf{86.8} & \underline{90.6} & \textbf{88.3} & \underline{89.6} & \textbf{88.1} & 88.3 & 86.1 & 91.0 & \textbf{88.0} & 86.6 & 78.2 & 88.9 & 87.2 & 90.4 & \textbf{87.7} & 90.7 & 85.5 & 48.6 \\
\rowcolor{my_green}Qwen3-Omni-30B-Instruct & \underline{86.6} & 83.1 & 83.3 & 89.1 & \underline{86.2} & 88.9 & \textbf{88.3} & 89.1 & 86.8 & 87.1 & 85.6 & 88.7 & \underline{87.4} & 73.8 & 77.7 & 87.8 & 86.8 & \textbf{91.0} & 87.3 & \textbf{90.9} & 85.5 & 77.4 \\
\rowcolor{my_green}MCAT-27B & 86.1 & 84.8 & 85.2 & \underline{90.0} & 85.9 & 90.1 & 87.2 & 89.5 & 86.5 & \underline{88.7} & \underline{86.9} & \underline{91.3} & 86.3 & \underline{87.2} & 78.9 & 89.2 & 87.6 & 90.2 & 86.8 & 90.4 & 88.2 & 79.7 \\ \midrule
\rowcolor{my_green}ESRT-1B (ours) & 83.1 & 82.7 & 83.5 & 87.3 & 84.0 & 86.7 & 85.5 & 86.7 & 83.2 & 85.8 & 83.1 & 86.4 & 83.7 & 83.2 & 76.0 & 86.1 & 83.5 & 88.5 & 84.6 & 88.4 & 86.0 & 79.9 \\
\rowcolor{my_green}ESRT-4B (ours) & 85.4 & \underline{86.4} & \underline{85.3} & 89.5 & 86.0 & 89.8 & \underline{87.3} & 89.2 & 86.3 & 88.5 & 86.8 & 90.8 & 85.8 & 87.0 & 78.6 & \underline{89.5} & 87.7 & 89.9 & 86.8 & 90.0 & \underline{89.1} & \textbf{83.2} \\
\rowcolor{my_green}ESRT-12B (ours) & \textbf{86.7} & \textbf{86.8} & \textbf{86.5} & \textbf{90.3} & \textbf{86.8} & \textbf{90.7} & \textbf{88.3} & \textbf{90.2} & \underline{87.2} & \textbf{89.3} & \textbf{88.0} & \textbf{91.8} & 87.1 & \textbf{88.0} & \textbf{80.1} & \textbf{90.2} & \textbf{88.9} & \underline{90.5} & \underline{87.6} & \underline{90.8} & \textbf{89.8} & \underline{83.0} \\
\midrule
\rowcolor{gray!8}
 & & & & & & & & & & & & & & & & & & & & & & & \multicolumn{1}{|c}{} \\
\rowcolor{gray!8}
\multirow{-2}{*}{\textbf{Models}} & \multirow{-2}{*}{kor} & \multirow{-2}{*}{lao} & \multirow{-2}{*}{msa} & \multirow{-2}{*}{mya} & \multirow{-2}{*}{nld} & \multirow{-2}{*}{nob} & \multirow{-2}{*}{pol} & \multirow{-2}{*}{por} & \multirow{-2}{*}{ron} & \multirow{-2}{*}{rus} & \multirow{-2}{*}{slk} & \multirow{-2}{*}{slv} & \multirow{-2}{*}{spa} & \multirow{-2}{*}{swe} & \multirow{-2}{*}{tam} & \multirow{-2}{*}{tha} & \multirow{-2}{*}{tgl} & \multirow{-2}{*}{tur} & \multirow{-2}{*}{urd} & \multirow{-2}{*}{uzb} & \multirow{-2}{*}{vie} & \multirow{-2}{*}{yue} & \multicolumn{1}{|c}{\multirow{-2}{*}{\textbf{Avg.}}} \\
\midrule
\rowcolor{my_green}Whisper+NLLB-3.3B & 86.7 & 79.3 & 86.5 & 79.8 & 83.7 & 86.2 & 85.3 & 86.4 & 86.9 & 86.0 & 86.2 & 85.1 & 83.7 & 88.0 & 84.4 & 80.6 & 82.0 & 86.8 & 78.6 & 87.7 & 85.6 & 76.1 & \multicolumn{1}{|c}{84.3} \\
\rowcolor{my_green}Gemini-3.5-Flash-Lite & 88.4 & 81.4 & 87.6 & 87.1 & 85.5 & 87.9 & 87.1 & 87.3 & 88.5 & 87.8 & 88.4 & 87.8 & 84.2 & 88.7 & 87.6 & \textbf{87.1} & 82.9 & \underline{88.3} & \underline{81.2} & 88.2 & 87.1 & 81.7 & \multicolumn{1}{|c}{86.6} \\
\rowcolor{my_green}SeamlessM4T-V2-Large & 85.1 & 81.4 & 86.6 & 85.7 & 85.1 & 86.8 & 85.7 & 86.6 & 87.8 & 86.3 & 87.3 & 86.8 & 83.2 & 88.4 & 87.3 & 82.1 & 83.3 & 86.7 & 79.4 & 87.7 & 85.6 & 79.8 & \multicolumn{1}{|c}{85.3} \\
\rowcolor{my_green}ZeroSwot-Large & 75.5 & 76.8 & 83.6 & 83.4 & 82.1 & 79.4 & 80.4 & 80.0 & 83.5 & 83.2 & 71.2 & 84.0 & 74.5 & 86.0 & 86.0 & 77.0 & 73.2 & 84.4 & 78.0 & 82.9 & 82.8 & 79.2 & \multicolumn{1}{|c}{80.2} \\
\rowcolor{my_green}Qwen2.5-Omni-7B & 84.8 & 39.0 & 83.3 & 41.4 & 82.3 & 83.4 & 81.3 & 86.6 & 81.3 & 85.2 & 76.6 & 70.6 & 83.4 & 84.1 & 50.4 & 55.9 & 82.0 & 79.4 & 51.4 & 47.0 & 76.0 & 84.4 & \multicolumn{1}{|c}{73.4}\\
\rowcolor{my_green}Gemma4-12B-it & 86.0 & 78.9 & 86.6 & 80.0 & 83.7 & 87.0 & 85.2 & 85.7 & 85.0 & 83.2 & 81.4 & 79.7 & 82.5 & 87.2 & 84.5 & 82.1 & 81.7 & 85.6 & 76.6 & 84.7 & 83.2 & 82.7 & \multicolumn{1}{|c}{83.3} \\
\rowcolor{my_green}Voxtral-24B & \underline{89.2} & 64.1 & 88.0 & 72.7 & \underline{87.1} & \underline{89.3} & \underline{88.4} & \textbf{89.1} & \underline{89.3} & \underline{89.1} & 87.6 & 87.3 & \textbf{86.0} & \underline{89.9} & 86.0 & \underline{86.7} & 81.8 & 87.9 & 78.0 & 87.3 & \underline{88.3} & \underline{88.0} & \multicolumn{1}{|c}{85.6} \\
\rowcolor{my_green}Qwen3-Omni-30B-Instruct & \textbf{89.7} & 80.3 & \underline{88.2} & 71.9 & 86.0 & 88.4 & 87.3 & 88.5 & 88.8 & 88.9 & 87.1 & 85.5 & 85.4 & 88.7 & 85.3 & 80.1 & \textbf{88.9} & 88.1 & 78.3 & 79.7 & \textbf{88.6} & \textbf{88.6} & \multicolumn{1}{|c}{85.7} \\
\rowcolor{my_green}MCAT-27B & 88.5 & 83.1 & 87.5 & 85.0 & 86.6 & 88.7 & 88.1 & 87.9 & 89.1 & 88.8 & 88.8 & \underline{88.5} & 85.2 & 89.3 & \underline{88.3} & 83.3 & 87.7 & 88.2 & 80.9 & 88.2 & 87.8 & \underline{88.0} & \multicolumn{1}{|c}{87.1} \\ \midrule
\rowcolor{my_green}ESRT-1B (ours) & 86.3 & 80.5 & 86.5 & 84.5 & 84.0 & 86.8 & 84.2 & 85.8 & 86.1 & 85.9 & 85.0 & 84.5 & 82.8 & 86.8 & 85.7 & 85.0 & 81.6 & 85.3 & 78.6 & 86.2 & 86.1 & 85.7 & \multicolumn{1}{|c}{84.6} \\
\rowcolor{my_green}ESRT-4B (ours) & 88.5 & \underline{84.1} & 87.8 & \underline{87.6} & 86.3 & 88.4 & 87.6 & 87.6 & 89.0 & 88.0 & \underline{89.0} & 88.1 & 84.8 & 89.1 & 87.8 & 83.1 & 87.4 & 87.9 & 81.1 & \underline{89.0} & 87.7 & 87.5 & \multicolumn{1}{|c}{\underline{87.2}} \\
\rowcolor{my_green}ESRT-12B (ours) & 88.9 & \textbf{84.5} & \textbf{88.6} & \textbf{88.8} & \textbf{87.2} & \textbf{89.5} & \textbf{88.6} & \underline{88.6} & \textbf{89.8} & \textbf{89.4} & \textbf{89.9} & \textbf{89.5} & \underline{85.5} & \textbf{90.0} & \textbf{88.9} & 84.1 & \underline{88.3} & \textbf{89.0} & \textbf{82.5} & \textbf{89.7} & \underline{88.3} & \textbf{88.6} & \multicolumn{1}{|c}{\textbf{88.1}} \\
\bottomrule
\end{tabular}}
\raggedright{\hspace*{1em}\textbf{Bold} values indicate the best results, while \underline{underlined} values denote the second-best results.}
\label{tab:eng44_comet}
\end{table*}
\endgroup

\subsection{Eng$\to$X S2TT on FLEURS}

\subsubsection{\textbf{Overall Performance}}
As shown in Table~\ref{tab:eng44_comet}, ESRT-12B achieves the highest average COMET score of \textbf{88.1} over all 44 English-to-X directions. It exceeds the strongest prior model, MCAT-27B, by 1.0 point despite using fewer than half as many parameters. The margins over Voxtral-24B, Qwen3-Omni-30B, Gemini-3.5-Flash-Lite, and SeamlessM4T-V2-Large are 2.5, 2.4, 1.5, and 2.8 points, respectively. ESRT-12B also outperforms the Whisper--NLLB cascade by 3.8 points. Thus, its advantage holds across specialized S2TT systems, general-purpose MLLMs, API models, and cascaded pipelines rather than depending on one baseline category. The consistently high average establishes a stronger performance ceiling with broad multilingual coverage.

\subsubsection{\textbf{Direction-Wise Dominance}}
Beyond the average, ESRT-12B obtains the best score in 31 of the 44 directions, demonstrating that its advantage is broadly distributed rather than concentrated in a few favorable targets. It exceeds COMET 88 on 29 directions, with particularly high scores for Finnish (91.8), Japanese (90.8), Indonesian (90.5), and Bulgarian (90.3). These gains span different scripts and language families. Even on difficult directions, ESRT-12B remains robust, reaching 80.1 on Hindi, 82.5 on Urdu, and 83.0 on Khmer.

\subsubsection{\textbf{Low-Resource Language Robustness}}
The largest gaps arise on languages with limited speech resources. ESRT-12B obtains 88.8 COMET on Burmese, 84.5 on Lao, and 83.0 on Khmer, whereas Qwen2.5-Omni-7B falls to 41.4, 39.0, and 38.0 and Voxtral-24B reaches only 72.7, 64.1, and 48.6. Its 89.8 on Kazakh and 89.7 on Uzbek further show that balanced multilingual training effectively reduces high-resource bias.

\subsubsection{\textbf{Cascaded and MLLM Baselines}}
The Whisper--NLLB cascade averages 84.3 COMET, trailing ESRT-12B by 3.8 points as recognition and translation errors accumulate. Large MLLMs improve the mean but remain uneven: Qwen3-Omni drops to 71.9 on Burmese, while Voxtral reaches 90.7 on Japanese but only 48.6 on Khmer. ESRT avoids these sharp collapses through specialized speech representations and balanced multilingual supervision.

\subsubsection{\textbf{Scaling and Parameter Efficiency}}
Average COMET increases from 84.6 for ESRT-1B to 87.2 for ESRT-4B and 88.1 for ESRT-12B. ESRT-4B already surpasses MCAT-27B and exceeds Voxtral-24B and Qwen3-Omni-30B by 1.6 and 1.5 points with far fewer parameters. Scaling to 12B improves 43 of 44 directions; only Khmer changes marginally from 83.2 to 83.0. This near-monotonic gain demonstrates an efficient accuracy--size trade-off for edge scenarios.

\newpage

\newpage

\clearpage

\begin{table*}[t]
  \centering
  \caption{COMET Results of Ablation Studies on the FLEURS Dataset.}
  \label{tab:ablation_refined}
  \renewcommand{\arraystretch}{1.0} 
  \setlength{\tabcolsep}{4.5pt} 
  \resizebox{1.0\textwidth}{!}{
  \begin{tabular}{lcccccccccccccccc}
  \toprule 
  \textbf{eng $\rightarrow$ X} & {heb} & {khm} & {lao} & {mya} & {yue} & {ind} & {kor} & {rus} & {tha} & {tur} & {ara} & {cmn} & {deu} & {fra} & {jpn} & \textbf{Avg.} \\ 
  \midrule
  ESRT-4B & 87.0 & 83.2 & 84.1 & 87.6 & 87.5 & 89.9 & 88.5 & 88.0 & 87.4 & 87.9 & 85.4 & 87.3 & 86.3 & 85.8 & 90.0 & 87.1 \\
  \hspace{1em}- Stage I & 66.6 & 68.6 & 68.9 & 76.2 & 67.6 & 68.0 & 70.1 & 67.7 & 67.9 & 67.7 & 67.3 & 65.4 & 63.2 & 60.3 & 72.4 & 67.9 {\color{red}(-19.2)} \\
  \hspace{1em}- Stage II & 86.9 & 83.0 & 84.0 & 87.5 & 87.4 & 89.8 & 88.4 & 87.9 & 87.3 & 87.8 & 85.3 & 87.1 & 86.2 & 85.6 & 89.9 & 86.9 {\color{red}(-0.2)} \\
  \hspace{1em}- Stage III & 65.1 & 64.8 & 65.8 & 52.0 & 72.0 & 73.2 & 79.0 & 62.3 & 75.2 & 77.8 & 61.8 & 62.4 & 80.0 & 70.3 & 74.4 & 69.1 {\color{red}(-18.0)} \\
  \hspace{1em}- LLM LoRA & 86.2 & 81.7 & 82.8 & 87.5 & 86.9 & 89.2 & 87.9 & 87.6 & 86.9 & 87.6 & 85.3 & 87.0 & 86.0 & 84.9 & 90.0 & 86.5 {\color{red}(-0.6)} \\
  \hspace{1em}+ Beam size 5 & 87.9 & 84.1 & 85.2 & 88.6 & 88.0 & 90.3 & 88.8 & 89.0 & 87.9 & 88.6 & 86.5 & 87.6 & 87.1 & 86.6 & 90.6 & 87.8 {\color{acl_green}(+0.7)} \\
  \bottomrule
  \end{tabular}}
\end{table*}

\subsection{Ablation Study}

\begin{figure}[t]
 \centering 
    \includegraphics[width=\linewidth]{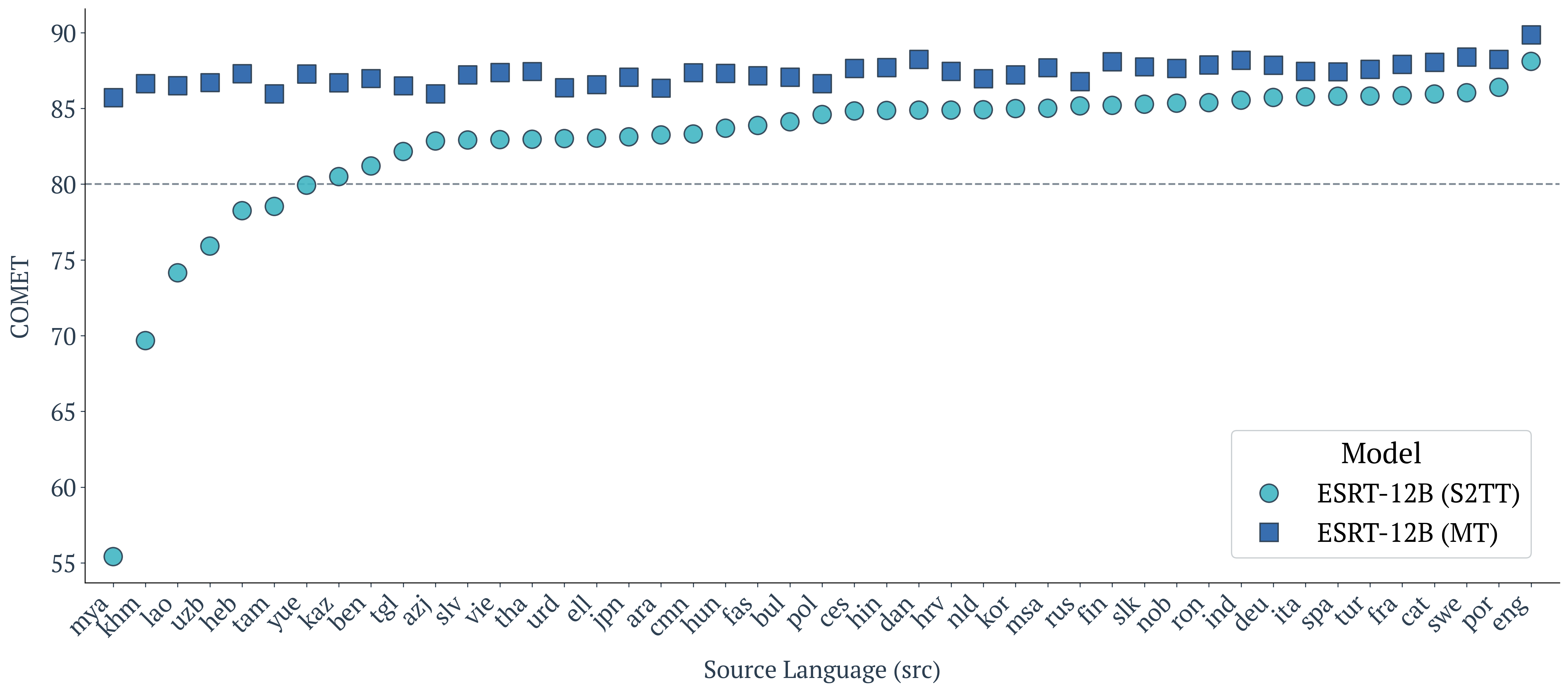}
\caption{\textbf{COMET Scores Between MT and S2TT.} The results show a
positive association between MT and S2TT performance across the evaluated
languages; this correlation does not by itself establish a causal relationship.}
    \label{fig:s2tt_mt}
\end{figure}

\subsubsection{\textbf{Multi-Task Weighted Curriculum Learning}}
Table~\ref{tab:ablation_refined} validates the ordering and weights of the three-stage curriculum. \textbf{Stage I} establishes acoustic grounding through ASR before translation is introduced. Removing it causes the largest drop, from 87.1 to 67.9 COMET ($-19.2$), showing that later supervision cannot recover content without a stable speech--text mapping. \textbf{Stage II} mixes ASR, speech-guided MT, and SRT with weights of 0.2/0.4/0.4. Although its removal changes the endpoint by only 0.2 point, it bridges transcription and translation, limits forgetting, and accelerates convergence. \textbf{Stage III} shifts the ASR/SRT weights to 0.2/0.8, activating direct speech translation while preserving recognition. Omitting it lowers COMET to 69.1 ($-18.0$) and Burmese to 52.0, indicating that intermediate supervision cannot replace end-to-end SRT training. Thus, Stage I builds the acoustic foundation, Stage II aligns the modalities, and Stage III enables cross-lingual generation.

\subsubsection{\textbf{Frozen LLM vs. LLM LoRA Training}}
Freezing the LLM backbone (\textit{- LLM LoRA}) decreases the average from 87.1 to 86.5 COMET. The small loss shows that the pretrained LLM already contains strong translation knowledge and that the adapter plus curriculum can map speech into its semantic space. LoRA still improves most directions, especially Khmer and Lao, by adapting the decoder to speech-conditioned representations. Low-rank updates therefore provide useful refinement without the cost of full-model fine-tuning.

\subsubsection{\textbf{Beam Search vs. Greedy Search}}
Beam search with width five (\textit{+ Beam Search (beam size 5)}) raises the average from 87.1 to 87.8 COMET, with gains to 84.1 on Khmer and 85.2 on Lao. Alternative hypotheses mitigate early errors. Its smaller gain shows that decoding only complements training.

\newpage

\subsection{Discussion}
\begin{table}[t]
  \centering
  \small
  \caption{Effect of Additional Training Data.}
  \label{tab:covost2}
  \renewcommand{\arraystretch}{1.1} 
  \setlength{\tabcolsep}{6pt} 
  \resizebox{\columnwidth}{!}{%
  \begin{tabular}{l ccc ccc} 
    \toprule 
    \multirow{2.5}{*}{Model} & \multicolumn{2}{c}{Training Data} & \multicolumn{4}{c}{CoVoST-2 (Test)} \\ 
    \cmidrule(lr){2-3} \cmidrule(lr){4-7}
    & Source & S2TT (h) & cmn & deu & jpn & \textbf{Avg.} \\ 
    \midrule 
    ESRT-4B    & FLEURS   & 7.5   & 82.7 & 80.8 & 85.6 & 83.0 \\
    ESRT-4B* & CoVoST-2 & 429.6 & 85.1 & 83.4 & 87.3 & 85.3 \\ 
    \bottomrule
  \end{tabular}}

  \vspace{2pt}
  \raggedright\footnotesize * indicates the same architecture with different training data.
\end{table}

\subsubsection{\textbf{MT vs. S2TT Analysis}}
As shown in Figure~\ref{fig:s2tt_mt}, we compared its MT and S2TT performance across 45 languages.
The results show a positive association between MT and S2TT performance across the evaluated languages, suggesting that stronger text-translation capability may be related to stronger speech-translation performance. Across the 45 languages, the model achieves a score above 80 in 38 languages.

\begin{table}[t]
\centering
\caption{Performance Comparison Between M5 and 5880ada.}
\label{tab:hardware_row_comparison_slim}
\renewcommand{\arraystretch}{1} 
\resizebox{\columnwidth}{!}{%
\begin{tabular}{llcccc}
\toprule
\multirow{2.5}{*}{\textbf{Hardware}} & \multirow{2.5}{*}{\textbf{Stage}} & \multirow{2.5}{*}{\textbf{Batch}} & \multicolumn{2}{c}{\textbf{Memory (GB)}} & \multirow{2.5}{*}{\makecell{\textbf{Speed}\\\textbf{(it/s)}}} \\
\cmidrule(lr){4-5}
 & & & \textbf{Model} & \textbf{Total} & \\
\midrule
\multirow{3}{*}{\makecell{Apple\\M5}}        & Cache Build    & 1  & 2.6 & $<4^{\dagger}$  & 1.9 \\
                                             & LLM Inference  & 1  & 8.6 & $<16^{\dagger}$  & 0.2 \\
                                             & LLM Inference  & 16 & 8.6 & $<16^{\dagger}$  & 0.9 \\
\midrule
\multirow{4}{*}{\makecell{Nvidia\\5880ada}}  & Cache Build    & 1  & 2.6 & $<4^{\ddagger}$  & 32.4 \\
                                             & LLM Inference  & 1  & 8.6 & $<10^{\ddagger}$ & 0.4 \\
                                             & LLM Inference  & 16 & 8.6 & $<10^{\ddagger}$ & 4.9 \\
                                             & vLLM Inference & -- & 12.2 & $<16^{\ddagger}$ & 60.9 \\
\bottomrule
\multicolumn{6}{l}{\scriptsize $^{\dagger}$ System-wide Shared Unified Memory. $^{\ddagger}$ GPU VRAM allocation managed by CUDA.} \\
\end{tabular}
}
\end{table}

\subsubsection{\textbf{Effect of Training Data Scale}}
Table~\ref{tab:covost2} compares two ESRT-4B models with the same architecture but different training corpora. Replacing 7.5 hours of FLEURS supervision with 429.6 hours of in-domain CoVoST-2 supervision increases the average COMET score on the three evaluated CoVoST-2 directions from 83.0 to \textbf{85.3}. This result shows that the larger in-domain training set improves performance.

\subsubsection{\textbf{Deployment of ESRT-4B in Edge Scenarios}}
As shown in Table~\ref{tab:hardware_row_comparison_slim}, ESRT-4B is successfully deployed on consumer edge devices. Benefiting from a unified memory architecture, the LLM inference stage requires an 8.6~GB footprint. Scaling the batch size from 1 to 16 keeps memory usage safely under 16~GB while boosting throughput from 0.2 to 0.9~it/s. Furthermore, Edge--cloud split inference requires less than 4~GB of CPU memory on the edge.

\newpage

\begin{figure}[t]
  \centering 
     \includegraphics[width=\linewidth]{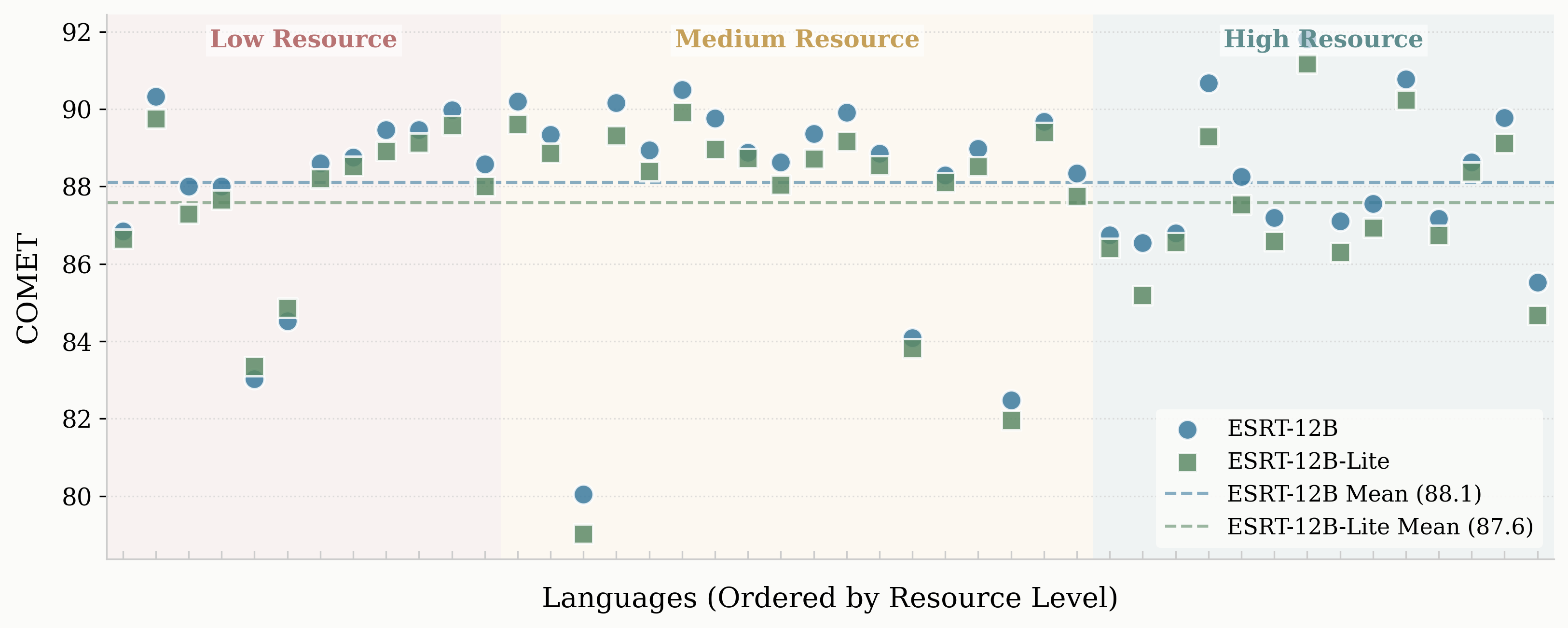}
 \caption{\textbf{Performance comparison across different speech-token budgets.} Results for ESRT-12B (80 speech tokens) versus ESRT-12B-Lite (40 speech tokens).}
     \label{fig:lite}
 \end{figure}

\begin{table}[t]
  \centering
  \small
  \caption{Raw-Audio-to-Representation Size Reduction.}
  \label{tab:bandwidth_fixed}
  \renewcommand{\arraystretch}{1.0} 
  \setlength{\tabcolsep}{4pt} 
  \resizebox{\columnwidth}{!}{%
  \begin{tabular}{lccc} \toprule
    \textbf{Model} & \textbf{Audio (MB)} & \textbf{Tensor (MB)} & \textbf{Size Reduction} \\ \midrule

    Cloud Api   & 392 & -- & $1.00\times$ \\ \midrule

    ESRT (ours)      & -- & 77 & $5.1\times$ \\
    ESRT-Lite (ours) & -- & 38 & $\mathbf{10.2\times}$ \\ \bottomrule
  \end{tabular}}
\end{table}

\subsubsection{\textbf{ESRT-12B vs. ESRT-12B-Lite}} 
We evaluate the impact of speech-token budgets by comparing ESRT-12B (80 speech tokens) with its compressed variant, ESRT-12B-Lite (40 speech tokens).
As shown in Figure~\ref{fig:lite}, ESRT-12B achieves the top average COMET score of \textbf{88.1}, while ESRT-12B-Lite maintains a strong \textbf{87.6}. 
Despite a substantial 50\% token reduction, the Lite version exhibits minimal performance degradation, demonstrating remarkable efficiency.

\subsubsection{\textbf{Practical Tensor-Size Analysis}}
Our evaluation set contains 647 audio samples occupying 392~MB. As reported in Table~\ref{tab:bandwidth_fixed}, the corresponding ESRT and ESRT-Lite tensors occupy 77~MB and 38~MB, yielding raw-audio-to-tensor size reductions of $\mathbf{5.1\times}$ and $\mathbf{10.2\times}$, respectively. These ratios exclude serialization, protocol overhead, and returned text; they quantify tensor size rather than measured network traffic.

\subsubsection{\textbf{Speech Reconstruction}}
\begin{figure}[h]
 \centering
    \includegraphics[width=\linewidth]{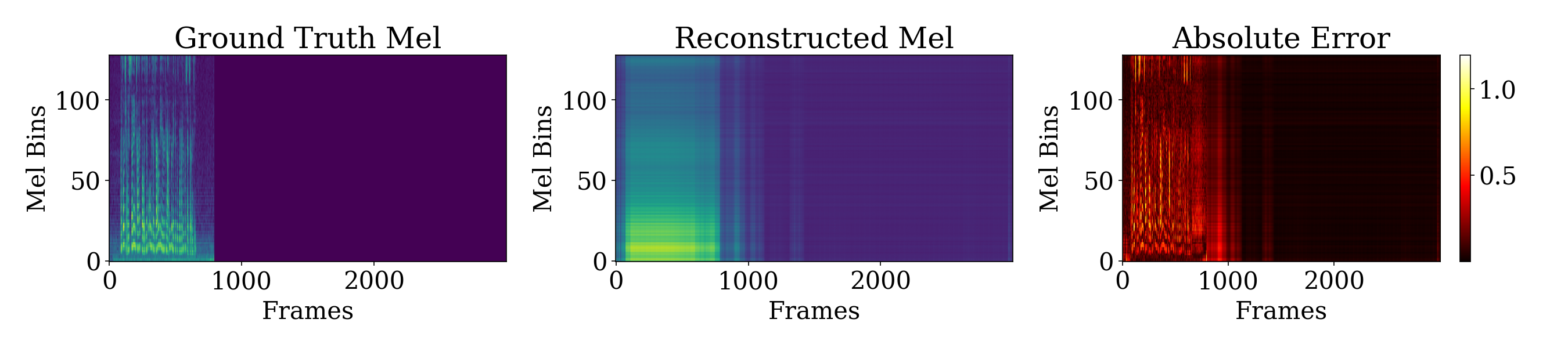}
\caption{\textbf{Speech reconstruction.} The image-inpainting reconstruction maps $(40,768)$ tensors to $(128,3000)$ features. The recovered audio remains highly noisy and unintelligible.}
    \label{error}
\end{figure}

Edge--cloud communication uploads a compressed tensor rather than the original waveform. To examine whether speech can be recovered directly in one attack setting, we train an image-inpainting network to predict acoustic features from the transmitted 40-query tensor. Figure~\ref{error} shows that the resulting reconstruction captures coarse temporal structure and approximate duration but remains noisy and unintelligible in this example. This qualitative result suggests that direct reconstruction is difficult for the attacker.

\section{Conclusion}
We presented ESRT, a parameter- and bandwidth-efficient Edge--cloud framework for many-to-many S2TT. Its weighted curriculum enables robust translation across 45 FLEURS languages, while split inference keeps raw speech on-device and reduces the transmitted tensor size by $5.1\times$. The reconstruction example provides preliminary evidence that intelligible speech is difficult to recover under the evaluated setting. Future work will improve low-resource coverage and evaluate information leakage more systematically.

\section*{Limitations}
ESRT inherits the 30-second input limit and low-resource weaknesses of Whisper, while its language coverage depends on the LLM. It is privacy-aware rather than formally privacy-preserving: intermediate representations may still expose linguistic or speaker information. Stronger inversion and attribute-inference attacks remain future work.

\bibliographystyle{IEEEtran}
\bibliography{custom} 

\vfill

\end{document}